\pdfoutput=1

\documentclass[11pt]{article}

\usepackage[]{ACL2023}

\usepackage{times}
\usepackage{latexsym}

\usepackage[T1]{fontenc}

\usepackage[utf8]{inputenc}

\usepackage{microtype}

\usepackage{inconsolata}

\usepackage{amsthm}
\usepackage{amsmath}
\theoremstyle{definition}
\newtheorem{definition}{Definition}[section]

\usepackage{graphicx}
\usepackage{tabularx}

\title{Training Data Extraction From Pre-trained Language Models: A Survey}

\author{Shotaro Ishihara \\
  Nikkei Inc. \\
  1-3-7, Otemachi, Chiyoda-ku, Tokyo \\
  \texttt{shotaro.ishihara@nex.nikkei.com} \\}

\begin{document}
\maketitle
\begin{abstract}
As the deployment of pre-trained language models (PLMs) expands, pressing security concerns have arisen regarding the potential for malicious extraction of training data, posing a threat to data privacy.
This study is the first to provide a comprehensive survey of training data extraction from PLMs.
Our review covers more than 100 key papers in fields such as natural language processing and security.
First, preliminary knowledge is recapped and a taxonomy of various definitions of memorization is presented.
The approaches for attack and defense are then systemized.
Furthermore, the empirical findings of several quantitative studies are highlighted.
Finally, future research directions based on this review are suggested.
\end{abstract}

\section{Introduction}

Pre-trained language models (PLMs) are widely used in natural language processing.
Statistical models that assign probabilities to token sequences have been studied, and large neural networks are increasingly being used for pre-training with large datasets.
This scaling has led to fluent natural language generation and success in many other downstream tasks~\citep{devlin-etal-2019-bert}.
In some cases, parameter updates are not required for downstream tasks~\citep{radford2019language,Brown2020-cf}.

With increasing applications of PLMs, security concerns have increased considerably~\citep{Bender2021-ki,Bommasani2021-sc,Weidinger2022-nn}.
Studies have revealed the risk of language models exhibiting unintentional \textit{memorization} of training data, and occasionally outputting memorized information~\citep{Carlini2019-pf,Carlini2021-xi,Carlini2022-wv,10.1145/3543507.3583199}.
In particular, \citet{Carlini2021-xi} identified that personal information can be extracted by generating numerous sentences from PLMs and performing \textit{membership inference}~\citep{Shokri2017-tr}.
These attacks on PLMs are referred to as \textit{training data extraction} and are undesirable because of privacy, decreased utility, and reduced fairness concerns~\citep{Carlini2022-wv}.
However, with the evolution of PLMs, limited progress has been achieved in addressing these concerns, and security technology is yet to mature.

This study is the first to provide a comprehensive survey of training data extraction from PLMs.
Starting with the pioneering work, we reviewed more than 100 previous and subsequent studies.
Specifically, we screened papers citing \citet{Carlini2021-xi}\footnote{\url{https://scholar.google.com/scholar?cites=12274731957504198296}} based on the relationships, the number of citations, and their acceptance.
First, Section~\ref{sec:Preliminaries} presents preliminary knowledge.
We then discuss several topics with the following contributions:
\begin{itemize}
  \setlength{\parskip}{0cm}
  \item A taxonomy of various \textbf{definitions of memorization} (Section~\ref{sec:Memorization}) was presented. Training data extraction has become close to the famous security attack known as model inversion~\citep{Fredrikson2015-fm}.
  \item We systematize the approaches to \textbf{attack} (Section~\ref{sec:Attacks}) and \textbf{defense} (Section~\ref{sec:Defenses}). Furthermore, we highlight \textbf{empirical findings} (Section \ref{sec:Evaluation}) from several quantitative evaluation studies.
  \item Based on the review, we suggest \textbf{future research directions} (Section~\ref{sec:Conclusion}).
\end{itemize}

\section{Preliminaries about PLMs}
\label{sec:Preliminaries}

This section describes the basics of modern PLMs.
First, we explain the methodology used for training language models and generating texts.
Next, the standard practical schema is introduced.

\subsection{Language Models}
\label{subsec:lm}

Language models represent a probability distribution over the sequences of tokens.
Based on the pre-training method, language modeling can be categorized into two types~\citep{Yang2023-go}: \textit{autoregressive language modeling}, which predicts words sequentially from left to right~\citep{Bengio2000-av,Mikolov2010-op}, and \textit{masked language modeling}, which hides some parts of a sentence and fills in the gaps~\citep{devlin-etal-2019-bert}.
The former is sometimes called \textit{causal language modeling}~\citep{Tirumala2022-qs}.

This study is focused on autoregressive language models with transformer ~\citep{NIPS2017_3f5ee243}, following many recent studies on training data extraction.
Note that some studies have focused on masked language models such as BERT~\citep{lehman-etal-2021-bert,mireshghallah-etal-2022-quantifying,he-etal-2022-extracted} and T5~\citep{Carlini2022-wv}.
Most studies address pre-training rather than fine-tuning~\citep{mireshghallah-etal-2022-empirical}.

Autoregressive language models take a series of tokens as input and output a probability distribution for the next token.
We show a schema of training and generation by following \citet{Carlini2021-xi}.

\paragraph{Training.}

The following statistical model was assumed for distribution:
\setlength{\abovedisplayskip}{4pt}
\setlength{\belowdisplayskip}{4pt}
\[\textbf{Pr}(x_1, x_2, \ldots, x_n),\]
where $x_1, x_2, \ldots, x_n$ is a sequence of tokens from a vocabulary using the chain rule of probability:
\[\textbf{Pr}(x_1, x_2, \dots, x_n) = \Pi_{i=1}^n \textbf{Pr}(x_i \mid x_1, \dots, x_{i-1}).\]
Let $f_\theta(x_i \mid x_1, \dots, x_{i-1})$ denote the likelihood of token $x_i$ when evaluating neural network $f$ with parameters $\theta$.
Language models are trained to optimize the probability of the data in a training set.
Formally, training involves minimizing the loss function as follows:
\[\mathcal{L}(\theta) = -\log \Pi_{i=1}^n f_\theta(x_i \mid x_1, \dots, x_{i-1})\]
for each data in the training set.
This setting can be qualitatively regarded as memorizing the flow of sentences in each training data.

\paragraph{Generating.}

New tokens can be generated by iterating the following process:
\begin{enumerate}
  \setlength{\parskip}{0cm}
  \setlength{\itemsep}{0cm}
  \item Choose $\hat{x}_{i+1} \sim f_\theta(x_{i+1} | x_1, \dots, x_i)$.
  \item Feed $\hat{x}_{i+1}$ back into the model to choose $\hat{x}_{i+2} \sim f_\theta(x_{i+2} | x_1, \dots, \hat{x}_{i+1})$.
\end{enumerate}
This decoding process continues until conditions are satisfied.
The simplest is greedy decoding, selecting the most probable tokens one by one.
However, studies have revealed that simply maximizing the output probability generates text that is not natural to humans~\citep{li-etal-2016-diversity,Holtzman2020-ne}.
Therefore, several approaches have been proposed for sampling from a probability distribution such as top-k sampling~\citep{fan-etal-2018-hierarchical} and top-p sampling (Appendix~\ref{sec:decoding}).

\subsection{Pre-training and Fine-tuning}
\label{subsec:plm}

Prior to BERT~\citep{devlin-etal-2019-bert}, specific models were trained for individual tasks.
By contrast, in the PLMs approach, large neural networks with large datasets are pre-trained and fine-tuned for several downstream tasks.
\citet{Alec2018} revealed that autoregressive language modeling is effective for PLMs with transformers.
This extension, GPT-2 \citep{radford2019language} and GPT-3 \citep{Brown2020-cf}, can be applied to various tasks without fine-tuning by providing a few examples (in-context learning).
The scaling of large models with large datasets has attracted considerable research attention (Appendix~\ref{sec:scaling}).

PLMs exhibit a significant advantage in using datasets that match a specific domain.
These models can exhibit superior performance in domain-specific tasks than larger models pre-trained on general datasets.
Studies, such as {{B}io{M}egatron}~\citep{shin-etal-2020-biomegatron}, {BioGPT}~\citep{Luo2022-am}, {Galactica}~\citep{Taylor2022-ep}, and {BloombergGPT}~\citep{Wu2023-og}, have been conducted.
However, the potential risk of training data extraction, especially when using sensitive datasets in pre-training, should be considered~\citep{Nakamura2020-pg,lehman-etal-2021-bert,Jagannatha2021-pf,Singhal2022-dq,Yang2022-ct}.
There are also ethical topics such as the human rights in the texts~\citep{Li2018-ft,Ginart2019-jd,Garg2020-as,Henderson2022-xv} and plagiarism regarding copyright~\citep{10.1145/3543507.3583199}.
Examples include PLMs created from contracts~\citep{chalkidis-etal-2020-legal,Zheng2021-vg}, clinical information~\citep{Kawazoe2021-do}, music~\citep{Agostinelli2023-ko}, and source code~\citep{Chen2021-cg}.

\section{Definitions of Memorization}
\label{sec:Memorization}

\begin{figure}[t]
  \centering
  \includegraphics[width=6.5cm]{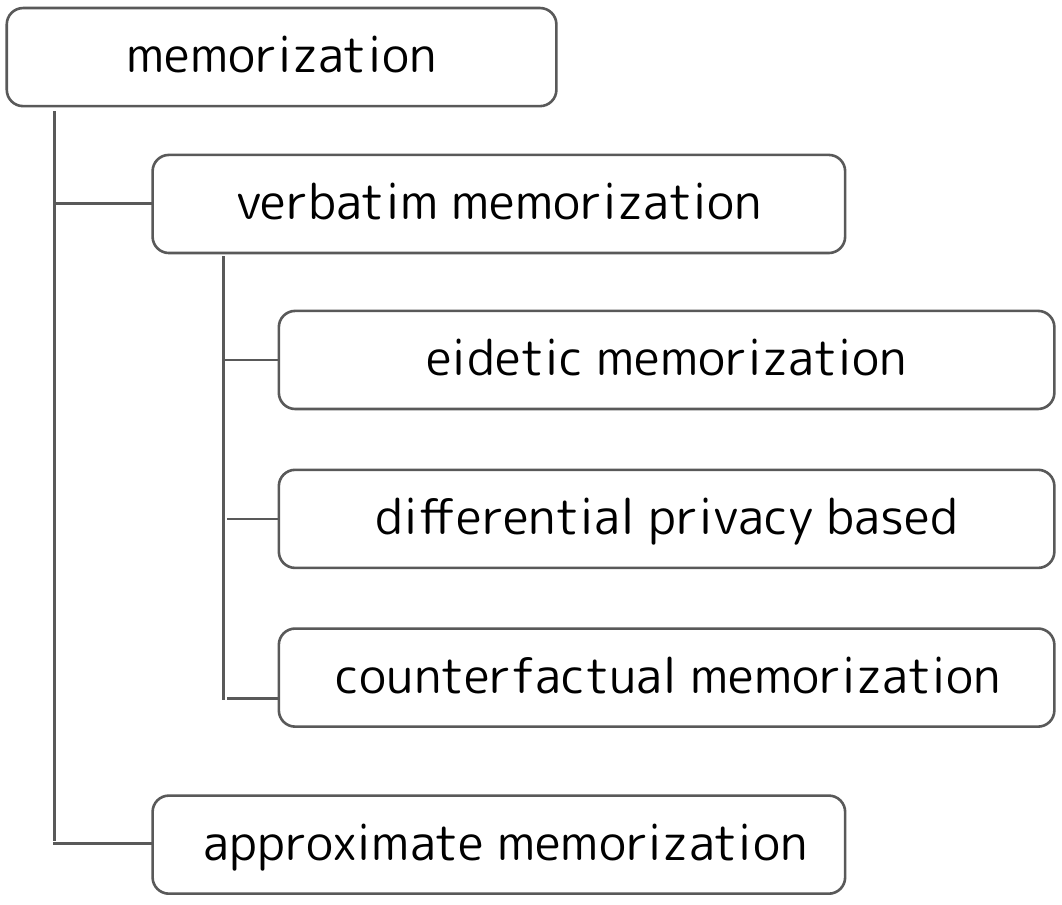}
  \caption{
    Taxonomy of definitions of memorization.
  }
  \label{fig:def_memorization}
\end{figure}

Memorization is the concept that PLMs store and output information about the training data.
There is a wide variety of research on memorization, with diverse definitions and assumptions.
We illustrate a taxonomy of definitions in Figure~\ref{fig:def_memorization}.

\subsection{Eidetic memorization}

A mainstream method is \textit{eidetic memorization} \citep{Carlini2021-xi} and its variations \citep{Thomas_McCoy2021-yq,Carlini2022-wv,pmlr-v162-kandpal22a,Tirumala2022-qs}.
These definitions assume that PLMs output memorized data when appropriate prompts are provided.
\citet{Carlini2021-xi} defined eidetic memorization as Definition \ref{def:eidetic_memorization}, and in a subsequent study~\citep{Carlini2022-wv}, they adopted the definition in Definition \ref{def:memorization}.
They stated that eidetic memorization can be used in cases in which no prompt, whereas the subsequent definition is suitable for conditions with prompts.
Some studies have adopted definitions similar to those in Definition \ref{def:memorization}.
Examples include~\citet{Tirumala2022-qs} with a per-token definition of \textit{exact memorization}, and \citet{pmlr-v162-kandpal22a} with a document-level definition of \textit{perfect memorization}.
\begin{definition}[eidetic memorization]
\label{def:eidetic_memorization}
A string $s$ is $k$-eidetic memorized by PLM $f_\theta$ if 
a prompt $p$ exists such that $f(p)=s$ and $s$ appears at most $k$ times in the training set.
\end{definition}
\begin{definition}[a variation of eidetic memorization]
\label{def:memorization}
A string $s$ is $k$-memorized with $k$ tokens of context from a PLM $f_\theta$ if a (length-$k$) string $p$ exists such that the concatenation $[p || s]$ is contained in the training set, and $f_\theta$ produces $s$ when prompted with $p$ by using greedy decoding.
\end{definition}

\subsection{Differential privacy}

Differential privacy~\citep{Dwork2006-jr} is widely used in memorization, and definitions based on differential privacy have been devised~\citep{Jagielski2020-wl,Nasr2021-qu}.
Differential privacy was formulated based on the premise that removing any data from the training set should not considerably change trained models.
Although this method protects the personal information of a single user, \citet{Brown2022-wl} reported that the method cannot capture the complexity of social and linguistic data.
Differential privacy is introduced as a defense approach in Section~\ref{subsec:TrainingDP}.

\subsection{Counterfactual memorization}
\label{subsec:Counterfactual}

Studies have defined \textit{counterfactual memorization} as the difference between a training data’s expected loss under a model that has and has not been trained on that data~\citep{Feldman2020-qi,Van_den_Burg2021-di}.
\citet{Zhang2021-sh} investigated this form of memorization in PLMs based on the taxonomy of human memorization in psychology.

The definition of counterfactual memorization has received limited attention in training data extraction.
\citet{Carlini2022-wv} noted that this definition requires training thousands of models to measure privacy.
Thus, evaluating PLMs becomes difficult because of their inference costs.
Furthermore, \citet{pmlr-v162-kandpal22a} remarked that this definition is not considered a privacy attack scenario because access to the training corpus is assumed.
This phenomenon is related to the adversarial knowledge presented in Section~\ref{subsec:membership}.

\subsection{Approximate memorization}
\label{subsec:Approximate}

\begin{figure*}[ht]
  \centering
  \includegraphics[width=14.5cm]{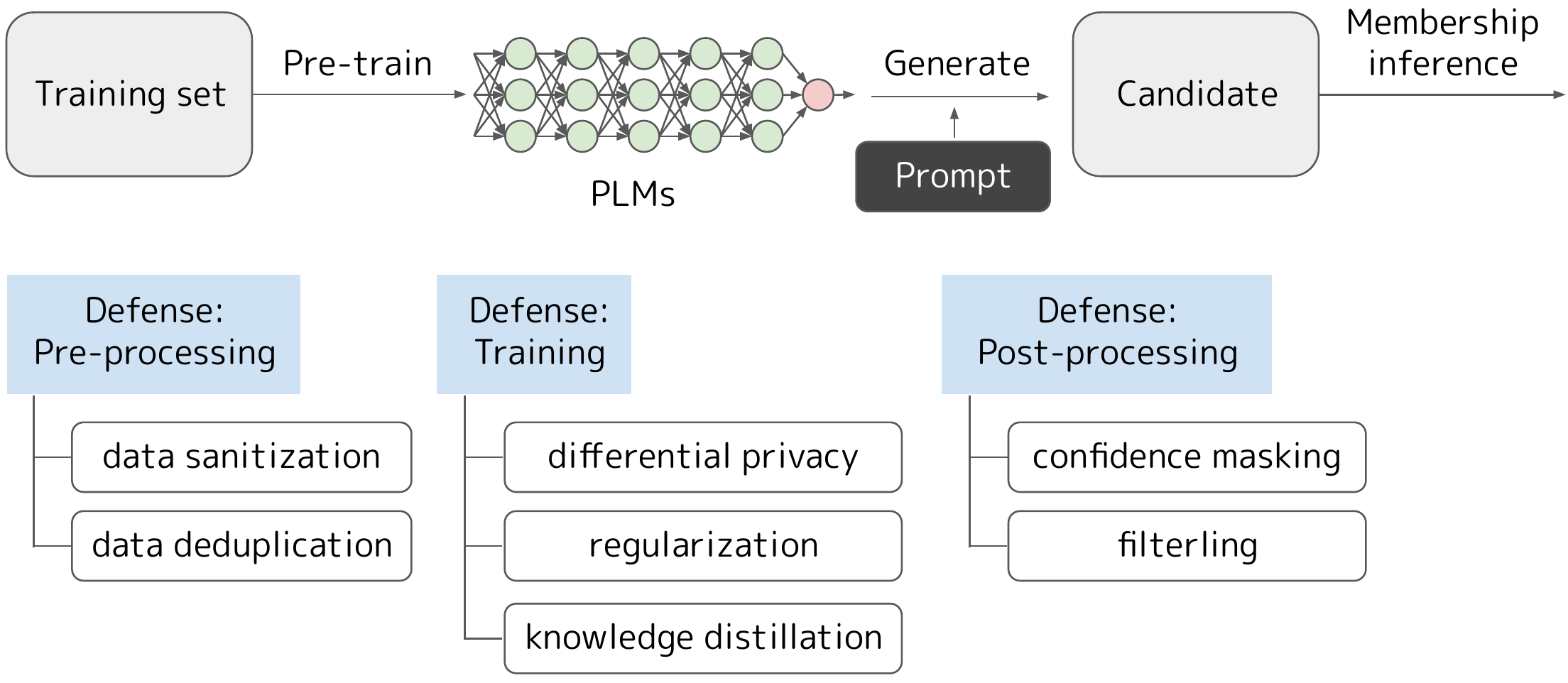}
  \caption{
    The procedure of training data extraction attacks and possible defenses.
  }
  \label{fig:overview}
\end{figure*}

Although the definitions of memorization thus far assume exact string matches, definitions have been proposed to relax this condition.
Here, \citet{Ippolito2022-dn} refer to definitions based on exact string matches as \textit{verbatim memorization}.
They revealed that verbatim memorization can be handled by simply adjusting the decoding method and proposed alternative definitions called \textit{approximate memorization} that consider string fuzziness, as presented in Definition \ref{def:approximate_memorization}.
Some methods have been proposed to calculate similarity.
\citet{Ippolito2022-dn} set the condition that $\rm{BLEU}(s, g)$~\citep{papineni-etal-2002-bleu} is greater than 0.75.
The threshold value of 0.75 was selected by qualitatively inspecting examples.
\citet{lee-etal-2022-deduplicating} defined that the token is memorized if it is part of a substring of 50 tokens of a string in the training data.
\begin{definition}[approximate memorization]
\label{def:approximate_memorization}
A string $s$ is $k$-approximately memorized by PLM $f_\theta$ if a (length-$k$) string $p$ exists such that $(s, g)$ satisfies certain conditions of similarity, and $f_\theta$ produces $g$ when prompted with p.
\end{definition}

\subsection{Revisiting model inversion}

Reconstructing training data from a model presents a well-known security concern called model inversion attacks~\citep{Fredrikson2015-fm}.
\citet{Carlini2021-xi} explained that the main difference is that training data extraction does not allow fuzziness.
However, this difference has decreased since the introduction of relaxed definitions of memorization.
\citet{pmlr-v162-kandpal22a} mentioned several previous studies~\citep{Carlini2019-pf,Carlini2021-xi,Inan2021-so} as model inversion.

\section{Training Data Extraction Attacks}
\label{sec:Attacks}

This section systematizes the attack procedure.
Most studies follow \citet{Carlini2021-xi}.
They revealed that hundreds of verbatim text sequences can be extracted from the training data.
Given a PLM, the procedure consists of two steps, candidate generation, and membership inference, as displayed in Figure~\ref{fig:overview}.

\subsection{Candidate generation}
\label{subsec:Candidate}

The first step is to generate numerous texts from a given PLM.
Texts can be generated from PLMs using several decoding methods, as discussed in Appendix~\ref{sec:decoding}.
Here, \citet{Carlini2022-wv} reported that the choice of the decoding strategy does not considerably affect their experimental results.
In contrast, \citet{10.1145/3543507.3583199} observed that top-k and top-p sampling tended to extract more training data.

Another perspective is the procedure for providing prompts.
Prompts are provided according to two options, giving only a special token\footnote{\citet{Carlini2021-xi} used \texttt{<|endoftext|>}, as indicated at \url{https://github.com/ftramer/LM_Memorization}.} (sometimes called \textit{no prompt}) or specific strings as prompts.
Studies have constructed prompts by extracting data from the dataset considered to be used in creating PLMs.
\citet{Carlini2021-xi} randomly sampled between 5 and 10 tokens from scraped data.
\citet{Carlini2022-wv} extracted a subset of the Pile dataset~\citep{Gao2020-kc} in prompting GPT-Neo model family~\citep{black-etal-2022-gpt}.

\subsection{Membership inference}
\label{subsec:membership}

Membership inference aims to predict whether any particular example is used to train a machine learning model~\citep{Shokri2017-tr,Song2019-ih,hisamoto-etal-2020-membership}.
This result can lead directly to privacy violations.
We describe membership inference on PLMs from the following five perspectives in a survey paper~\citep{Hu2021-uj}: target model, adversarial knowledge, approach, algorithm, and domain.

\paragraph{Target model.}

This study focuses on autoregressive language models as discussed in Section~\ref{subsec:lm}.
Attacks on other models such as word embeddings~\citep{Song2020-cp,Mahloujifar2021-wa,meehan-etal-2022-sentence}, natural language understanding~\citep{parikh-etal-2022-canary}, text classification~\citep{Nasr2019-wu,Zhang2022-fl,elmahdy-etal-2022-privacy}, and image diffusion models~\citep{Carlini2023-yo} exist but are not covered.

\paragraph{Adversarial knowledge.}

The second perspective is the knowledge that can be handled explicitly by attackers.
We describe two aspects of adversarial knowledge, namely models and training sets.
The patterns of adversarial knowledge in this study are summarized in Appendix \ref{sec:adversarial-knowledge}.

\citet{Hu2021-uj} presented the adversarial knowledge of models.
The models are classified into two categories, namely white-box and black-box, according to accessibility~\citep{Nasr2019-wu}.
Under the white-box setting, an attacker can obtain all information and use it for the attack.
This includes the training procedure and the architecture and trained parameters of the target model.
However, in the black-box setting, an attacker can only have limited access to the target model.
\citet{Hu2021-uj} classified the black-box setting into three parts, namely full confidence scores, top-k confidence scores, and prediction labels only.
They differ in the extent of access an attacker has to the PLMs output.
The setting of full confidence scores assumes a situation in which the training process of the model is unknown, but all outputs for any given input are available.
Therefore, an attacker can obtain prediction labels with probabilities and calculate the loss.
The setting of top-k confidence scores indicates that an attacker can obtain several candidates of the output.
The scope of the attack is restricted because losses cannot be calculated.
Another setting provides only labels without prediction values~\citep{pmlr-v139-choquette-choo21a,10003239}.
Many web services with PLMs, such as DeepL\footnote{\url{https://www.deepl.com/translator}} and ChatGPT\footnote{\url{https://openai.com/blog/chatgpt/}}, only allow users to view labels for the model output.

Furthermore, we describe the adversarial knowledge of the training sets.
In the white-box setting, the training set is stated and publicly available.
The most harmful attacks are black box setups that do not assume access to the training set.
Such attacks include PLMs created by private datasets.
In some cases, the data are partially publicly available.
Such cases include the ones wherein only the beginning of the news article is available for free, certain editions are accessible, and some articles have been made private over time.
Although the data itself are not partially published, substrings can be inferred in the hidden private data using a priori knowledge \citep{Henderson2018-uv,Carlini2019-pf}.
Examples are prompts like \textit{"Bob’s phone number is"} and \textit{"Alice's password is"}.

We must be aware of scenarios in which the dataset and PLMs are unwillingly leaked and become public.
Adversarial knowledge is immediately converted to the white-box level.
For example, even if a web service with PLMs trained on a private dataset provides users with only a string, it is crucial to discuss risks when both the dataset and the PLMs are unintentionally made public.

\paragraph{Approach.}

\citet{Hu2021-uj} divided the membership inference approaches into three categories, namely classifier-based~\citep{Shokri2017-tr,Song2019-ih}, metric-based~\citep{Bentley2020-yu,pmlr-v139-choquette-choo21a,Song2021-gu}, and differential comparisons~\citep{HuiYYBGC21}.
For example, in shadow training~\citep{Shokri2017-tr,Song2019-ih}, a primary classifier-based method, additional training is assumed in the model (white-box settings).
Some metric-based methods can be applied to realistic black-box settings.

In studies of training data extraction from PLMs, \textit{perplexity} is often used for metrics of membership inference~\citep{Carlini2019-pf,Carlini2021-xi}. 
Given a sequence of tokens $x_1, \ldots, x_n$, the perplexity is defined as:
\[\mathcal{P} = \exp\left(-\frac{1}{n}\sum_{i = 1}^n\log f_\theta(x_i | x_1, \ldots, x_{i - 1})\right)\]

\paragraph{Algorithm.}

The fourth perspective is whether the algorithm is centralized or federated.
Federated learning approaches have received considerable attention in privacy protection research~\citep{Melis2019-ie,Nasr2019-wu,Lee2021-od,Kairouz2021-gv}.
However, focusing on training data extraction, the mainstream approach is based on centralized methods as of April 2023.

\paragraph{Domain.}

Text datasets are rooted in various domains, as described in Section~\ref{subsec:plm}.
Clinics are a crucial research field that involves handling of highly confidential information.
\citet{lehman-etal-2021-bert} recovered patient names and their associated conditions from PLMs using electronic clinical records.
\citet{Jagannatha2021-pf} demonstrated that patients with rare disease profiles may be highly vulnerable to higher privacy leakages through experiments using PLMs of clinical data.
Many other domains require careful processing, such as contracts~\citep{yin-habernal-2022-privacy} and source code\footnote{\url{https://github.blog/2021-06-30-github-copilot-research-recitation/}}.
A discussion of the right to be forgotten in the legal and news industries has emerged~\citep{Li2018-ft,Ginart2019-jd,Garg2020-as,Henderson2022-xv}.
Therefore, it should be ensured that PLMs do not unintentionally become digital archives.

Publicly available datasets do not necessarily indicate that they are completely independent of the risk of training data extraction from PLMs.
The context in which the information is shared should be known to respect privacy~\citep{Dourish2004-kh,Nissenbaum2009-dd}.
Nissenbaum’s contextual integrity~\citep{Nissenbaum2009-dd} states that a change in any one of five characteristics (data subject, sender, recipient, information type, and transmission principle) may alter privacy expectations.
\citet{Brown2022-wl} emphasized the importance of PLMs only with data explicitly intended for public use.
The Italian Data Protection Authority issued a statement\footnote{\url{https://www.garanteprivacy.it/home/docweb/-/docweb-display/docweb/9870847}} on March 2023 in accordance with the European General Data Protection Regulation (GDPR) against OpenAI, the provider of ChatGPT, for their data processing.

\section{Training Data Extraction Defenses}
\label{sec:Defenses}

This section systematizes approaches to defense.
We can mitigate privacy risks before, during, and after creating PLMs as displayed in Figure \ref{fig:overview}.
The classification was reconstructed using references~\citep{Hu2021-uj,huang-etal-2022-large,Jagielski2022-fp}.
Extensive studies have been conducted on the hazardous generation of PLMs~\citep{kurita-etal-2020-weight,mei-etal-2022-mitigating,levy-etal-2022-safetext,Ouyang2022-lr,Carlini2023-gk}.
However, this study focused on training data extraction.

\subsection{Pre-processing}

First, pre-processing the training set is considered.

\paragraph{Data sanitization.}

The simplest solution is to identify and remove any text that conveys personal information~\citep{Ren2016-rt,Continella2017-gz,vakili-etal-2022-downstream}.
However, as noted in Section~\ref{subsec:membership}, privacy depends on the context, and determining privacy from the string alone is difficult.
\citet{Brown2022-wl} proposed that data sanitization is only useful for removing context-independent, well-defined, static pieces of personal information from the training set.

\paragraph{Data deduplication.}

Studies have indicated that data deduplication mitigates the memorization of PLMs~\citep{Allamanis2019-mr,pmlr-v162-kandpal22a,lee-etal-2022-deduplicating}.
This method is more efficient than methods that train models and is expected to be a practical solution.
Empirical findings on data deduplication are presented in Section~\ref{subsec:deduplicate}.

\subsection{Training}
\label{subsec:TrainingDP}

The second method is a pre-training strategy.

\paragraph{Differential privacy.}

Applying differential privacy~\citep{Dwork2006-jr} methods for providing data privacy guarantees in machine learning models has attracted considerable research attention.
Differential privacy is a data protection measure that is designed to ensure that providing data does not reveal much information about the user.
However, applying these algorithms (e.g., DP-SGD~\citep{Abadi2016-jh} and DP-FedAvg~\citep{Ramaswamy2020-cd}) to PLMs is challenging.
Performance degradation and increased computation and memory usage are the primary concerns.

To address this problem, a framework has been proposed for training models in two steps~\citep{Yu2021-sb,Yu2021-uq,Li2021-ny,he2023exploring}\footnote{A study has also appeared that applies these algorithms to in-context learning settings~\citep{Panda2023-bw}.}.
In the framework, large amounts of non-private datasets are used for pre-training to obtain general features; next, additional training is applied with a sensitive dataset using a differential privacy algorithm.
\citet{Downey2022-ae} reported that the differential privacy approach is effective in preventing memorization, despite its computational and model performance costs.
Note that \citet{Tramer2022-ot} summarized a critical view.
They argued that publicly accessible datasets are not free from privacy risks because they contain information that is unintentionally released to the public.
Therefore, discussing whether private information that we want to hide is contained in the public dataset is essential.
It is known that understanding the semantic guarantee of differential privacy is difficult when private data is involved~\citep{Cummings2021-im}.

Another barrier to applying differential privacy to PLMs is the requirement of defining secret boundaries even though text data are not binary.
Studies have considered various levels of granularity, from individual tokens or words to sentences, documents, or even the entire user dataset \citep{Brendan_McMahan2017-zi,Levy2021-rq,lukas2023analyzing}.

\paragraph{Regularization.}

Regularization is a well-known approach for suppressing overfitting in machine learning models.
The memorization of models is typically associated with overfitting \citep{Yeom2018-ad,Zhang2021-tz}.
Therefore, regularization during training that reduces overfitting can be used as a measure of membership inference~\citep{Hu2021-uj}.
\citet{mireshghallah-etal-2021-privacy} proposed a regularization method regarding the memorization of PLMs and claimed usefulness compared with differential privacy methods.
Some studies have constrained the representation of neural networks by the information bottleneck layer~\citep{alemi2017deep,henderson2023a}.

Pre-training large neural networks has distinctive tendencies compared with common machine learning.
A single data in the training set is not used for too many epochs in pre-training and is sometimes used for less than one epoch.
Furthermore, \citet{Carlini2021-xi} reported that a characteristic of PLM memorization is the emergence of training data with an abnormally lower loss than the average.
\citet{Tirumala2022-qs} revealed that large language models can memorize most of their data before overfitting and tend not to forget much information through the training process.
\citet{Biderman2023-rj} have focused on the training process and attempted to predict the memorization of PLMs.

\paragraph{Knowledge distillation.}

Another approach is knowledge distillation \citep{Hinton2015-id}, in which the output of a large teacher model is used to train a small student model.
\citet{Shejwalkar2021-zt} revealed that knowledge distillation can be used to restrict an attacker’s direct access to a private training set, which considerably reduces membership information leakage.

\subsection{Post-processing}

The third step is to post-process the PLMs output.

\paragraph{Confidence masking.}

Limiting the output of PLMs is a simple but effective defense mechanism.
For example, confidence masking can be used for adjusting adversarial knowledge, as presented in Section~\ref{subsec:membership} and Appendix~\ref{sec:adversarial-knowledge}.

\paragraph{Filtering.}

Filtering the output of PLMs before providing them to users is crucial.
Identifying items to be filtered incurs a cost, and ensuring diversity remains challenging.
\citet{Perez2022-fe} proposed a method to automatically identify test cases by extracting potentially dangerous outputs by detailing prompts using various PLMs.

\section{Empirical Findings}
\label{sec:Evaluation}

This section presents empirical findings on training data extraction from PLMs.
Initial studies were limited to qualitative evaluations, but subsequent studies~\citep{lee-etal-2022-deduplicating,pmlr-v162-kandpal22a,Ippolito2022-dn,Tirumala2022-qs,Downey2022-ae,Carlini2022-wv,10.1145/3543507.3583199} have focused on quantitative evaluations.

In particular, based on one of the first comprehensive quantitative studies~\citep{Carlini2022-wv}, we report on the impact of the model size, the string duplication in the training set, and the length of prompts.
They used various sizes of GPT-Neo model family~\citep{black-etal-2022-gpt}, which are the autoregressive language models pre-trained by the Pile dataset~\citep{Gao2020-kc}.
Four model sizes, namely 125 million, 1.3 billion (B), 2.7 B, and 6 B parameters, were considered.
The number of duplicate strings was determined by analyzing the Pile dataset.
A subset of 50,000 sentences from the Pile dataset was used for evaluation, and the distribution of duplicates was considered.
The beginning of each sentence was cut out at a certain number of tokens and considered as a prompt.
The amount of memorization was calculated as the fraction of generations that exactly reproduce the true string for their prompt averaged over all prompts and sequence lengths.

\subsection{Larger models memorize more}

\citet{Carlini2022-wv} revealed that a near-perfect log-linear relationship exists such that the larger the model size is, the more strings are memorized.
Numerically, a ten-fold increase in the model size increased the amount of memorization by 19 ppt.
For comparison, they performed the same analysis with the GPT-2 model family.
The amount of memorization was 40 \% for 1.3 B GPT-neo compared with 6 \% for the GPT-2 of the same size.
This phenomenon implied the effect of memorization of the training data, not just the model size.

\citet{Carlini2022-wv} used the definition of verbatim memorization, and \citet{Ippolito2022-dn} confirmed similar results with the definition of approximate memorization.
Although not sufficiently quantitative, initial studies~\citep {Carlini2019-pf,Zhang2021-tz} have provided preliminary evidence.
\citet{Tirumala2022-qs} and \citet{10.1145/3543507.3583199} also revealed that larger models memorize more.

\subsection{Duplicate strings are memorized}
\label{subsec:deduplicate}

\citet{Carlini2022-wv} reported that a clear log-linear trend exists between the number of duplicates and the amount of memorization.
They measured the amount of memorization for each bucket with duplicate counts ranging from 2 to 900.
\citet{pmlr-v162-kandpal22a} and \citet{lee-etal-2022-deduplicating} also revealed that duplication in the training set of PLMs relates to the likelihood of memorizing strings and proposed that deduplication mitigates training data extraction.
However, memorization can occur even with only a few duplicates, and deduplication cannot prevent it completely.
\citet{Chang2023-ww} reported that the degree of memorization of ChatGPT and GPT-4~\citep{OpenAI2023-as} was related to the frequency of the passages that appeared on the web.

\subsection{Longer prompts extract more}

\citet{Carlini2022-wv} revealed that the amount of memorization increases with the length of the prompt.
For example, the amount of memorization by the 6 B model was 33 \% for 50 tokens, compared with 65 \% for 450 tokens.
This experiment was inspired by the findings of \citet {Carlini2019-pf}.
They suggested that setting the maximum prompt length available to users considerably reduces the risk of training data extraction.

\section{Conclusion \& Future Directions}
\label{sec:Conclusion}

We have reviewed over 100 papers for the first comprehensive survey on training data extraction from PLMs.
The final section provides suggestions for future research directions.
We hope that this study highlights the importance of training data extraction from PLMs and accelerates the discussion.

\subsection{Is memorization always evil?}

Most studies did not distinguish the degree of danger of memorized strings~\citep{Lee2020-pl}.
Ideally, the undesirable memorization of telephone numbers and email addresses must be separated from the acceptable memorization.
\citet{huang-etal-2022-large} was among the first to differentiate between memorization and association in PLMs.
They concluded that the risk of specific personal information being leaked is low because PLMs cannot semantically associate personal information with their owners.

The boundary between memorization and knowledge of PLMs remains ambiguous with the definition of approximate memorization~\citep{Ippolito2022-dn,lee-etal-2022-deduplicating}.
Deduplication of training sets, which is considered useful in Sections~\ref{sec:Defenses} and \ref{sec:Evaluation}, leads to the elimination of helpful knowledge.
Therefore, we must consider what memorization is~\citep{Haviv2022-uz} and balance the security concerns with the model performance, depending on the final application.
The definition of counterfactual memorization introduced in Section~\ref{subsec:Counterfactual} incorporated psychological findings that could be useful despite its challenges.

\subsection{Toward broader research fields}
\label{subsec:fuzziness}

Discussing the handling of the fuzziness of a string is important.
\citet{Ippolito2022-dn} stated that the current definition of approximate memorization focuses on English, and different considerations are required for other conditions such as non-English languages.
In addition, they suggested two research areas that could help improve the definition: image generation memorization and plagiarism detection.
Images are more difficult to generate than text for matching exactly with the original.
Therefore, fuzzy memorization has been investigated and measured.
\citet{Fredrikson2015-fm}, which proposed the model inversion attack, used face recognition in images as the subject of their experiments.
Studies have used metrics that consider image similarity~\citep{Zhang2020-yb,Haim2022-mk,Balle2022-lv}.
Furthermore, the trend toward pre-training in both images and language~\citep{NEURIPS2019_c74d97b0,Li_Duan_Fang_Gong_Jiang_2020} should be considered.
The limitations of the definition of verbatim textual matching have been discussed in plagiarism detection research ~\citep{Roy2009-jw,potthast-etal-2010-evaluation}.
Similarities are explored from multiple perspectives, including word changes, shuffling, and paraphrasing.

\subsection{Evaluation schema}

Room for ingenuity exists in the construction of evaluation sets.
Establishing a schema for quantitative evaluation, which has received considerable attention, is critical.
Studies mentioned in Sections \ref{sec:Attacks} and \ref{sec:Evaluation} have created evaluation sets by extracting a subset of the training set.
Sampling is essential because of inference time limitations.
However, we must be careful to see if there are other factors to consider besides the distribution of the number of duplicates to avoid bias due to sampling.

Evaluation metrics for the training data extraction are open for discussion.
\citet{Carlini2021-xz} postulated that the ideal evaluation metric must be based on realistic attack scenarios, whereas most studies on membership inference measure the average accuracy rate.
They proposed that membership inference should be evaluated by the true positive rate with a low false positive rate.
The Training Data Extraction Challenge\footnote{\url{https://github.com/google-research/lm-extraction-benchmark}} measures attack speed as well as recall and precision.

\clearpage
\section*{Limitations}

First, this study focused on PLMs in training data extraction, particularly autoregressive language models. 
Other target models, such as masked language models (described in Section~\ref{subsec:lm}) and word embeddings (noted in Section~\ref{subsec:membership}), require another discussion.
Additionally, due to prioritization constraints, the discussion on other topics, including model inversion attacks and the federated learning approach, was limited.
However, these areas are established and can be supplemented by other studies ~\citep{Fredrikson2015-fm,Zhang2021-es}.

Second, in practical applications of PLM, it is necessary to audit not only security but also various other aspects such as performance degradation~\citep{Mokander2023-jm}.
There are a number of security concerns beyond training data extraction (noted in Section~\ref{sec:Defenses}).
There are also papers discussing performance degradation of PLMs over time~\citep{ishihara-etal-2022-semantic}.

Finally, this comprehensive survey is based on information as of April 2023.
Studies on training data extraction from PLMs have primarily focused on natural language processing and security.
These domains are undergoing rapid changes.
Therefore, some of the content may become obsolete in the near future.

\section*{Ethics Statement}

The privacy concerns regarding training data extraction from PLMs were reviewed to help mature discussions in academia and industry.
Of course, its purpose is not to promote these attacks.

Studies on PLMs tend to focus on the English language, which is the language used by the majority of people in the world, and the same is true for training data extraction.
Therefore, this study focused on English.
As indicated in Section~\ref{subsec:fuzziness}, research on other languages is encouraged.

\section*{Acknowledgements}

We would like to thank Editage (\url{www.editage.com}) for English language editing.

\bibliographystyle{acl_natbib}

\begin{table*}[ht]
\centering
\begin{tabular}{lll}
\hline
\begin{tabular}[c]{@{}l@{}}Adversarial knowledge\end{tabular} & \multicolumn{1}{l}{Model or the output} & \multicolumn{1}{l}{Pattern} \\ \hline
white-box & all & Models are available with proper explanations. \\
black-box & full confidence scores & All outputs of models are available. \\
 & top-k confidence scores & Top-k outputs of models are available. \\
 & prediction label only & Only prediction labels are available. \\ \hline
\end{tabular}
\caption{
    Adversarial knowledge of models and patterns.
}
\label{tab:adversarial-knowledge-models}
\end{table*}

\begin{table*}[ht]
\centering
\begin{tabular}{lll}
\hline
\begin{tabular}[c]{@{}l@{}}Adversarial knowledge\end{tabular} & \multicolumn{1}{l}{Training set} & \multicolumn{1}{l}{Pattern} \\ \hline
white-box & all & Dataset used for training is stated and publicly available. \\
black-box & partial & Dataset used for training is stated but not available. \\
 &  & Dataset used for training is stated and partially available. \\
 & nothing & Dataset used for training is not stated. \\ \hline
\end{tabular}
\caption{
    Adversarial knowledge of training sets and patterns.
}
\label{tab:adversarial-knowledge-corpus}
\end{table*}

\appendix

\section{Type of Decoding}
\label{sec:decoding}

Two classes of methods, namely deterministic and stochastic, are used for decoding~\citep{Su2022-jj}.
In the deterministic method, the most probable tokens based on the probability distribution of the model are used.
Greedy methods and beam searches are widely used.
However, studies have revealed that simply maximizing the output probability generates text that is not natural to humans~\citep{li-etal-2016-diversity,Holtzman2020-ne}.
Therefore, several approaches have been proposed for sampling from a probability distribution.
Stochastic methods include top-k sampling~\citep{fan-etal-2018-hierarchical}, top-p sampling, and nucleus sampling~\citep{Holtzman2020-ne}, in which samples are extracted from the lexical subset.
A method to adjust the probability distribution using the temperature parameter was used to increase the diversity of the generated texts~\citep{Ackley1985-ng}.

In the candidate generation step in Section~\ref{subsec:Candidate}, texts can be generated from PLMs using several decoding methods.
Some studies adopted a greedy method \citep{Carlini2022-wv}.
Others used top-k sampling~\citep{Carlini2021-xi,lee-etal-2022-deduplicating} and tuned the temperature~\citep{Carlini2021-xi} to increase the diversity of the generated texts.

\section{Scaling Law for Language Models}
\label{sec:scaling}

Building PLMs requires large datasets.
Studies have proposed models with larger parameters pre-trained with large datasets~\citep{Smith2022-ft,Chowdhery2022-jc}.
Experimental results revealed the existence of a scaling law~\citep{Kaplan2020-vr,Henighan2020-rp}.
This study suggested that the performance of language models using the transformer improves as the model size, dataset size, and amount of computation increase.
\citet{Villalobos2022-ud} cautioned that the data available for pre-training language models may be exhausted in the near future.

\section{Patterns of Adversarial Knowledge}
\label{sec:adversarial-knowledge}

Table \ref{tab:adversarial-knowledge-models} presents the patterns of adversarial knowledge of the models and Table \ref{tab:adversarial-knowledge-corpus} details the patterns of adversarial knowledge of the training set.
These tables provide specific patterns.
For example, white-box for models indicates PLMs published on platforms such as Hugging Face\footnote{\url{https://huggingface.co/models}} with training explanations, which can be downloaded.
As discussed in Section~\ref{subsec:membership}, two main types, namely white and black boxes, exist.
In black-box settings, several patterns depend on the situation.
Table \ref{tab:adversarial-knowledge-models} reveals the classification of the black-box proposed by \citet{Hu2021-uj}: full confidence scores, top-k confidence scores, and prediction labels.
In Table \ref{tab:adversarial-knowledge-corpus}, several possible patterns of adversarial knowledge are presented on training sets.

\end{document}